\documentclass[10pt,twocolumn,letterpaper]{article}

\usepackage{cvpr}
\usepackage{times}
\usepackage{epsfig}
\usepackage{graphicx}
\usepackage{amsmath}
\usepackage{amssymb}
\usepackage{multirow}
\usepackage[inline]{enumitem}
\usepackage{dsfont}
\usepackage[dvipsnames]{xcolor}
\usepackage{multirow}
\usepackage{pifont}
\usepackage{algorithm}
\usepackage{algpseudocode}

\usepackage{caption}
\usepackage{subcaption}
\usepackage{fix2col}
\usepackage{xcolor}
\usepackage{diagbox}
\usepackage{xspace}
\usepackage{tabularx,booktabs}
\usepackage{blindtext}

\usepackage[pagebackref=true,breaklinks=true,letterpaper=true,colorlinks,bookmarks=false]{hyperref}

\usepackage{pifont}
\newcommand{\cmark}{\ding{51}}%
\newcommand{\xmark}{\ding{55}}%

\definecolor{mypink1}{rgb}{0.858, 0.188, 0.478}
\definecolor{mypink2}{RGB}{219, 48, 122}
\definecolor{myorange}{RGB}{0, 120, 0}

\newcommand{\takaaki}[1]{\textbf{\textcolor{red}{[TAKAAKI: #1]}}}
\newcommand{\rongyu}[1]{\textbf{\textcolor{mypink1}{[YU: #1]}}}
\newcommand{\han}[1]{\textbf{\textcolor{blue}{[HAN: #1]}}}

\newcommand{\delete}[1]{}

\cvprfinalcopy
\def\cvprPaperID{2330} 

\title{FrankMocap: Fast Monocular 3D Hand and Body Motion Capture \\ 
      by Regression and Integration
}

\author{ Yu Rong\textsuperscript{1,3}
	\hspace{0.3in} Takaaki Shiratori\textsuperscript{2}
	\hspace{0.3in} Hanbyul Joo\textsuperscript{3}
	\vspace{5pt}
	\\
    \textsuperscript{1}{The Chinese University of Hong Kong}
	\hspace{0.3in} \textsuperscript{2}{Facebook Reality Labs}
	\hspace{0.3in} \textsuperscript{3}{Facebook AI Research} \\ 
}

\let\oldtwocolumn\twocolumn
\renewcommand\twocolumn[1][]{%
	\oldtwocolumn[{#1}{
		\begin{center}
			\includegraphics[width=\linewidth]{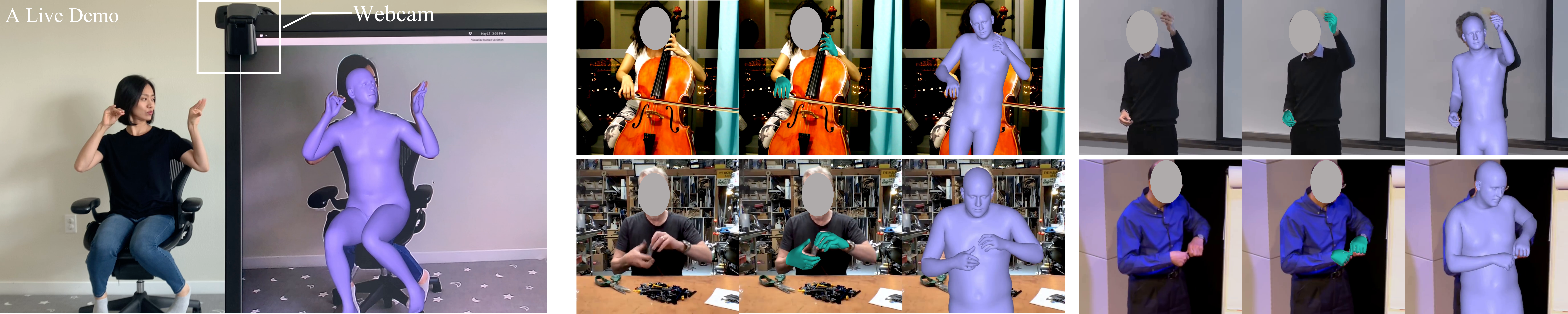}
			\captionof{figure}{\small 
			We present a whole 3D body motion capture system, named \emph{FrankMocap}, to simultaneously estimate 3D body and hand motion from monocular videos in the wild.
			Our system allows us to perform a live whole body motion capture demo using a single RGB webcam, as shown on the left. On the right, several example results on in-the-wild videos are demonstrated, where we show the input images (left), the 3D hand motion capture outputs (middle), and the whole body motion capture outputs (right).}
			\label{fig:teaser}
		\end{center}
	}]
}

\begin{document}
\maketitle

\begin{abstract}

Although the essential nuance of human motion is often conveyed as a combination of body movements and hand gestures, the existing monocular motion capture approaches mostly focus on either body motion capture only ignoring hand parts or hand motion capture only without considering body motion. 
In this paper, we present FrankMocap\footnote{``FrankMocap'' is an homage to the Frankenstein's monster in \emph{The Modern Prometheus.}}, a motion capture system that can estimate both 3D hand and body motion from in-the-wild monocular inputs with faster speed (9.5 fps) and better accuracy than previous work. 
%
%
%
To construct FrankMocap, we build the state-of-the-art monocular 3D ``hand'' motion capture method by taking the hand part of the whole body parametric model (SMPL-X). Our 3D hand motion capture output can be efficiently integrated to monocular body motion capture output, producing whole body motion results in a unified parrametric model structure.
We demonstrate the state-of-the-art performance of our hand motion capture system in public benchmarks, and show the high quality of our whole body motion capture result in various challenging real-world scenes, including a live demo scenario.
\footnote{Code and models are available at~\url{https://penincillin.github.io/frank_mocap}.}

\end{abstract}

\section{Introduction}
\label{sec:Introduction}

Billions of daily human activities are being recorded as videos and uploaded to public internet websites, capturing extremely diverse human behaviors in various real-world scenarios. 
%
%
A technology that can digitize the human motions from these videos has enormous potentials in various applications including human-computer interaction, social artificial intelligent, and robotics. 
A motion capture system with a commodity camera and reduced computation would enable people to make full use of such applications.

While the hands and body are equally important for human motion understanding toward these applications, 
the hands are physically small body parts, making it difficult to capture the motion of the hands and body jointly even with a professional motion capture system.
The same is true for recent work on 3D body pose (\ie, torso and limbs) estimation from a single RGB image~\cite{Bogo2016,kanazawa2018end,Xiang:2019:Monocular,kolotouros2019spin,kolotouros19convolutional}. 
Although the accuracy improvement of body pose estimation is significant, subtle finger gestures are ignored, losing the original nuance of the human motion. 
Similarly, there have been noticeable achievements in 3D hand pose estimation from a depth input~\cite{Oikonomidis-12,Sridhar-13,Sharp-15,Sridha-15,Tzionas-16,Ye-16} or a single RGB image~\cite{Zimmermann:2017:Learning, Cai_2018_ECCV, Iqbal_2018_ECCV, Boukhayma:2019:3D,Ge:2019:3D,Zhang:2019:End}. 
However, these approaches are often demonstrated with hand-specific camera views, rather than the more challenging in-the-wild scenarios where a camera is capturing whole bodies of people and the hands tend to be in low resolutions with frequent motion blurs.
A few recent approaches aim to capture 3D motions of the whole body (\ie, the body and hands) by leveraging the 3D parametric models that can express both hands and body~\cite{joo2018, Xiang:2019:Monocular, Pavlakos:2019:SMPLX}. However, these approaches rely on optimization techniques to fit the parametric models to image measurements, which are relatively slow and not suitable for real-time applications.
In this paper, we present a fast and accurate motion capture method to estimate both 3D body and hand poses from monocular RGB images or videos, as shown in Figure~\ref{fig:teaser}. 
Our method consists of two regression modules that predict 3D poses of the body and hands individually from a single RGB image input,
followed by an integration module that produces the whole body pose from the outputs from the body and hand modules
\footnote{We use \textit{whole body} to represent all the body parts including the fingers, while we use \textit{body} to represent the torso and limbs excluding the fingers.}.
A main idea of our approach is to make the outputs from body module and hand module as compatible as possible, enabling us to efficiently integrate the outputs for whole body motion capture. 
To make this integration process tractable, we employ the SMPL-X model~\cite{Pavlakos:2019:SMPLX} that has a unified skeleton and mesh representation for both the body and hands. 
Based on that, the body module and hand module contribute the different part of the same output structures.  
Inspired by the achievements of the recent work on 3D body motion estimation~\cite{kanazawa2018end, kolotouros2019spin}, the hand and body modules are designed based on deep neural network techniques, and directly regress 3D poses from an input single RGB image.
Given the 3D body and hand pose estimations from the hand and body modules, the integration can be performed by direct \textit{copy-and-paste} of the predicted body and hand poses to the SMPL-X model, achieving a near real-time performance for whole body 3D motion capture ($\sim$9.5 fps). 
A further improvement for better estimate of the whole body pose can be achieved via the optimization framework we also present. 
Follow the similar spirit of Joo~\etal~\cite{joo2018}, we call our method as FrankMocap to represent this regression and integration manner.

We demonstrate the fast and accurate performance of FrankMocap on various real-world monocular videos, including a real-time demo. Notably, our 3D hand pose estimation outperforms previous approaches in public benchmarks. We also present thorough ablation studies to demonstrate the advantage of our method, and compare our method with previous hand motion capture methods and whole body motion capture methods.

\begin{figure*}[t]
	\centering
	\includegraphics[width=0.95\linewidth]{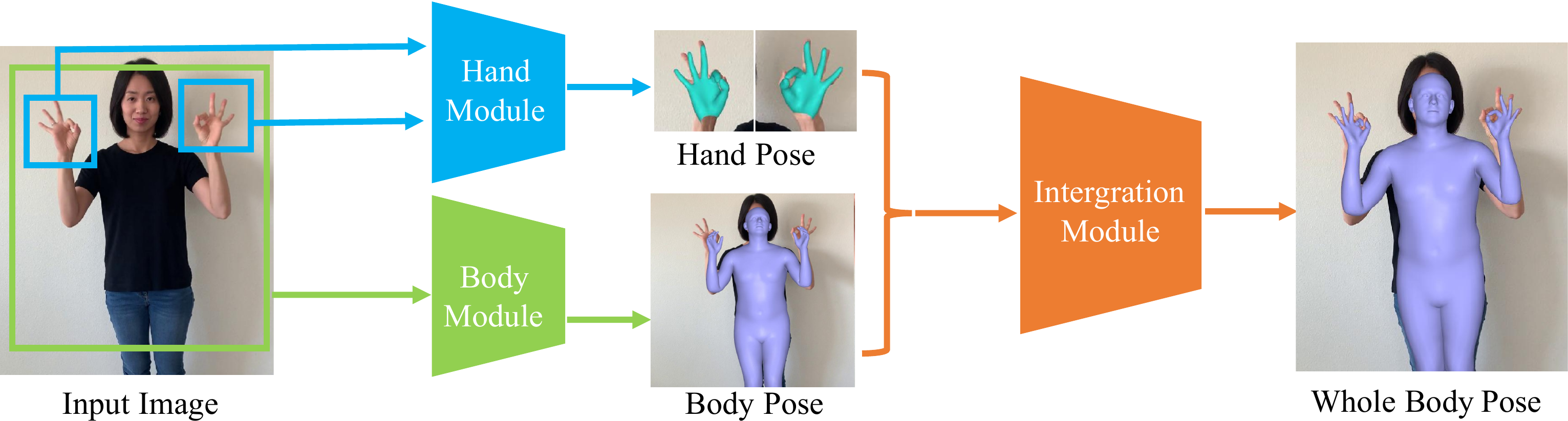}
	\caption{\small Overview of our pipeline for whole body motion capture. Given a single RGB image input, we apply our hand module and body module to estimate 3D hands and 3D body. Our integration module combines these outputs into a unified whole body output.}
	\label{fig:overview}
\end{figure*}
\section{Related Work}
\label{sec:RelatedWorks}

\noindent \textbf{3D Parametric Human Body Models.}
3D Parametric human body models are widely used for markerless motion capture, 
to model the deformation of 3D human (including body, face and hands) via low dimension parameters instead of the original vertices~\cite{anguelov2005scape,pons2015dyna, Pavlakos:2019:SMPLX, romero2017embodied, joo2018}. 
The SCAPE is a pioneering work that accounts for shape variations and pose deformations~\cite{anguelov2005scape}.
After that, Loper~\etal introduce the SMPL~\cite{Loper2015} that learns local pose-dependent blendshape on top of linear blend skinning for holistic mesh deformation as well as shape variations. 
Later, Romero~\etal~\cite{romero2017embodied} extend the SMPL to a hand and introduce the hand deformation model called MANO. They also develop a unified body and hand model called SMPL+H.
Joo~\etal build a unified model of body, hands and face, called Adam, and use it to achieve whole body motion capture of face, body and hands from a multiview setup~\cite{joo2018}.
Pavlakos~\etal~\cite{Pavlakos:2019:SMPLX} also similarly develop a unified model of body, hands and face, called SMPL-X, in which all template body parts are designed by artists for consistent quality over the body parts, and learned deformation statistics.

\noindent \textbf{Single Image 3D Body Pose Estimation.}
Many monocular 3D body pose estimation approaches consider to predict 3D body keypoint locations from single images~\cite{Ramakrishna2012, tan17indirect,Tung2017,martinez2017simple,pavlakos2017coarse}.
As the major limitations, the output of these methods cannot be directly used for graphics applications, since 3D joint angles are missing and the lengths of parts are not preserved. 
More recent monocular 3D body pose estimation approaches adopt parametric 3D human model such as  SMPL~\cite{Loper2015} or Adam model~\cite{joo2018,Xiang:2019:Monocular} for 3D body representation.
The use of 3D parametric models allows them to reconstruct a 3D body pose by fitting the 3D body model to 2D observations, such as 2D keypoints, via optimization framework~\cite{Bogo2016}. More recent work~\cite{kanazawa2018end,kolotouros2019spin,Tung2017, Rong2019, kolotouros19convolutional, Xu:19:denserac} leverages the deep learning framework to directly regress parameters of a body model. 

There are also approaches that use a hybrid framework by using a deep learning framework to produce an intermediate representation such as 2D and depth heat maps and fitting a skeletal model on these outputs to reconstruct joint angles~\cite{Mehta2017, Xiang:2019:Monocular, mehta2019xnect}.
%
Various types of inputs are considered in these deep network approaches, including single RGB images~\cite{kanazawa2018end}, 2D keypoint heatmaps~\cite{pavlakos18learning},
body part segmentation~\cite{omran18neural} or densepose maps~\cite{rong2019delving,Xu:19:denserac}.
Due to the lack of training data with 3D annotations, these models are trained with mixed datasets including indoor datasets such as Human3.6M~\cite{h36m_pami} and in-the-wild datasets such as COCO~\cite{lin2014microsoft} or 3DPW~\cite{vonMarcard2018}. 
Most papers in this area such as HMR~\cite{kanazawa2018end} and SPIN~\cite{kolotouros2019spin} use single images as input. 
There are some other works takes sequences as input. The representative ones are Zhang~\etal\cite{zhang2019predicting} and VIBE~\cite{kocabas2020vibe}.
In this paper, we mainly focus on processing single images and applying simple temporal smoothness to handle sequence inputs.

\noindent \textbf{Single Image 3D Hand Pose Estimation.}
%
Previous works on 3D hand joint estimation takes depth images as input~\cite{oikonomidis2011efficient,Oikonomidis-12,Sridhar-13,Sharp-15,Sridha-15,tagliasacchi2015robust,Tzionas-16,tkach2016sphere,Ye-16}. Although achieving good performance, these methods cannot be easily applied to in-the-wild RGB images and videos.
Recent work begins to use single RGB images as input~\cite{Zimmermann:2017:Learning,mueller2018ganerated,Cai_2018_ECCV,yang2019disentangling}. These approaches focus more on 3D hand joints location estimation instead of joint angles.
Inspired by recent success in 3D body motion capture~\cite{kanazawa2018end,kolotouros2019spin}, there are several methods on single image 3D hand pose estimation.
Boukhayma~\etal~\cite{Boukhayma:2019:3D} uses images and 2D pose predicted from OpenPose~\cite{Cao:2019:Openpose} as input and regress the parameters of the MANO model~\cite{romero2017embodied}.
Zhang~\etal~\cite{Zhang:2019:End} share a similar framework as Boukhayma. The key difference is that their model contains a 2D heatmap prediction module, instead of using predictions from OpenPose.
Baek~\etal's method~\cite{baek2019pushing} also has similar architecture. The difference lies in that they additionally adopt 2D masks as an intermediate representation.
Different from these approaches, the work of Ge~\etal~\cite{Ge:2019:3D} use a self-created 3D hand model instead of MANO. The proposed framework takes single images as input and predicts 2D heatmaps as intermediate representation. After that, graph convolutional network~\cite{kipf2017semi} is used to regress the vertices of the hand model.
%


\noindent \textbf{Joint 3D Pose Estimation of Body and Hands.}
There are a few methods ~\cite{Xiang:2019:Monocular,Pavlakos:2019:SMPLX} that pursue to estimate 3D poses of body and hands together.
Due to the lack of annotated data for whole body capture, all these previous works resort to optimization methods. 
SMPLify-X~\cite{Pavlakos:2019:SMPLX} uses the SMPL-X model~\cite{Pavlakos:2019:SMPLX} to represent a whole body pose. 
The model parameters are optimized by fitting to 2D keypoints with additional constraints including body pose priors and collision penalizer.
Monocular Total Capture (MTC) ~\cite{Xiang:2019:Monocular} is based on the Adam model~\cite{joo2018}. It adopts deep neural networks to get 2.5D predictions first. Then the parameters of Adam are obtained through optimization.
Both of these methods rely on optimization with relatively slow computation time (from 10 seconds to a few minutes).
Besides, when 2D heatmap detection failed, their accuracy degrades significantly.



 

%
%

\section{Method}
\label{sec:Method}

Our method aims to estimate 3D body (the torso and limb parts)
and 3D hands (both left and right) from monocular inputs (either monocular images or videos). 
Our method produces, as output, the parameters of the SMPL-X model~\cite{Pavlakos:2019:SMPLX} to represent both 3D body and hand poses in a unified form. 
An important aspect of our method is to use separate expert modules for each body and hand pose estimation while both modules produce the compatible outputs as part of SMPL-X model.
An overview of the framework is shown in Figure~\ref{fig:overview}.

Notably, our hand pose estimator leverages the hand part of the SMPL-X model, by treating it as a stand-alone parametric hand model.
While still showing the state-of-the-art monocular hand pose estimation performance, the output of our hand module can be directly merged to the body estimation output~\cite{kolotouros2019spin}, to pose the whole body SMPL-X model. 
We also present an optimization framework to improve the hand and body pose estimation output with additional constraints for better accuracy. 
In the remaining part of this section, we assume the inputs to our model are single images. Details of processing sequence inputs are included in section~\ref{sec:implementation_detail}.

%

\subsection{Overview of SMPL-X Model}

Given a single image input cropped around a single person, our method produces whole body motion capture output as a form of shape and pose parameters of the SMPL-X model. 
As an extension of the SMPL model~\cite{Loper2015}, the SMPL-X model can represent the shape variations and pose-dependent deformation of human bodies via a combination of low-dimensional shape and pose parameters. %
As a key difference from the SMPL model that only focuses on body parts, SMPL-X can also express finger motions and facial expressions, by including additional sets of parameters for them.

We formulate the SMPL-X model, denoted by $W$, as:

\begin{equation}
\label{eq:smplx}
	\boldsymbol{V}_w = W( \boldsymbol{\phi}_w, \boldsymbol{\theta}_w, \boldsymbol{\beta}_w), \\
\end{equation}

\noindent
where $W$ is parameterized by global orientation of the whole body $\boldsymbol{\phi}_w \in \mathbb{R}^{3}$, 
whole body pose parameters $\boldsymbol{\theta}_w  \in \mathbb{R}^{ (21 + 15 + 15) \times 3}$ accounting for pose-dependent deformation,
and shape parameters $\boldsymbol{\beta} \in \mathbb{R}^{10}$ accounting for cross-identity shape variations of the body and hands. 
We divide $\boldsymbol{\theta}_w$ for each of the body and hands, namely
body pose parameters $\boldsymbol{\theta}_w^b  \in \mathbb{R}^{21 \times 3}$, 
left hand pose parameters $\boldsymbol{\theta}_w^{lh} \in \mathbb{R}^{15 \times 3}$ (\ie, 3 joints per finger),
right hand pose parameters $\boldsymbol{\theta}_w^{rh} \in \mathbb{R}^{15 \times 3}$, and thus $\boldsymbol{\theta}_w = \{ \boldsymbol{\theta}_w^b, \boldsymbol{\theta}_w^{lh}, \boldsymbol{\theta}_w^{rh}  \}$.~\footnote{Note that we ignore other parameters of the original SMPL-X model, including facial expression parameters.}
All pose parameters are defined in the angle-axis representation which stores the relative rotation to the parent joints defined in the kinematics map. 
As output, the SMPL-X model produces a mesh structure with 10,745 vertices, $\boldsymbol{V}_w \in \mathbb{R}^{10475 \times 3}$. 
%
The 3D joint locations of the whole body can be obtained by applying a joint regression function $R$ from the posed vertices:
\begin{equation}
\boldsymbol{J}^{3D}_{w}  = R_w( \boldsymbol{V}_w),
\end{equation}
where $\boldsymbol{J}^{3D}_{w} \in \mathbb{R}^{ (22 + 15 +15) \times 3}$.

Our hand models are defined by taking the hand parts of SMPL-X:
\begin{equation}
\label{eq:smplx_hand}
	\boldsymbol{V}_h =  H(\boldsymbol{\phi}_h, \boldsymbol{\theta}_h, \boldsymbol{\beta}_h), \\
\end{equation}
\noindent
where $\boldsymbol{\theta}_h \in \mathbb{R}^{3 \times 15}$ is hand pose parameters and $\boldsymbol{\beta_h}$ is the shape parameters for hand model.
Since our hand model is taken from SMPL-X, $\boldsymbol{\beta_h}$ shares the same parameterization space as $\beta_w$. For hand model $H$, we only focus on its influence on the hand part.
For brevity, we use $\boldsymbol{\theta}_h$ to denote the hand pose parameters instead of $\boldsymbol{\theta}_{rh}$ or $\boldsymbol{\theta}_{lh}$ to describe our hand pose estimation method. 
%
%
$\boldsymbol{\phi}_h \in \mathbb{R}^{3}$ represents the global orientation of the hand meshes, which is necessary to use hand model as a stand-alone model, independent from the ancestry joints of the original SMPL-X model $W$. 
Our hand model $H$ produces the hand mesh structure with 778 vertices, $\boldsymbol{V}_h \in \mathbb{R}^{778 \times 3}$. Here, we define the hand mesh vertices from the whole body mesh $\boldsymbol{V}_w$ by cropping the vertices around the wrist area, where we choose all vertices from which any wrist joint and finger joints are closest. Since the hand mesh $\boldsymbol{V}_h$ is a subset of the whole body mesh $\boldsymbol{V}_w$, the exact vertex correspondences are known.
We also consider the 3D joint regression function for hand $\boldsymbol{J}^{3D}_{h}$:
\begin{equation}
\boldsymbol{J}^{3D}_{h}  = R_h(\boldsymbol{V}_h),
\end{equation}
where $\boldsymbol{J}^{3D}_{h} \in \mathbb{R}^{21 \times 3}$ contains a wrist, 15 finger joints (3 joints per finger), and 5 finger tips.
To define the joint regression matrix of $R_h$, we take the wrist and finger joint parts from the whole body joint regression matrix $R_w$, and define additional rows for the 5 finger tips that are not defined in $\boldsymbol{J}^{3D}_{w}$. See Figure~\ref{fig:hand_visualize} for visualization of $H$ and skeleton hierarchy. 

%
 
The major advantage of our representation is that the components of 3D hand model, including pose parameters, vertices, and 3D joints, are directly compatible with the whole body parameterization. This enables us to efficiently integrate outputs from the body module and the hand module. 



\begin{figure}[t]
	\begin{center}
		\includegraphics[width=\linewidth]{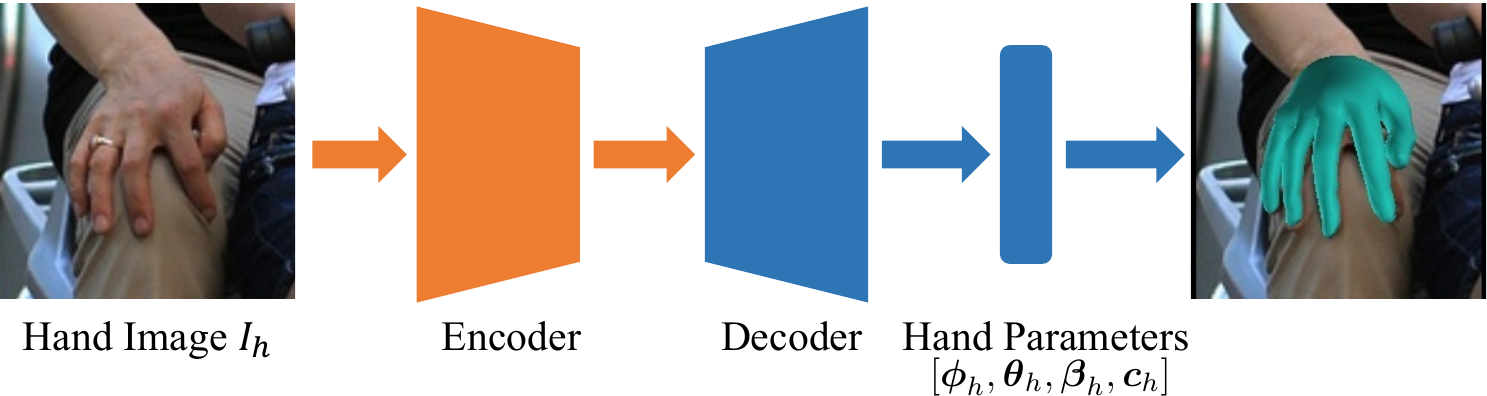}
	\end{center}
	\vskip -0.3cm
	\caption{\small \textbf{Overall framework of our hand module.} Our hand module takes a cropped hand image $\mathbf{I}_H$ as input, and produces the parameters of hand model, $[\boldsymbol{\phi}_h, \boldsymbol{\theta}_h, \boldsymbol{\beta}_h, \boldsymbol{c}_h] $. Our hand module is built by a deep encoder-decoder network. The predicted hand parameter is used to produce the mesh shape and pose of the hand part of SMPL-X.}
	\label{fig:hand_model}
	\vspace{0.1cm}
\end{figure}

\subsection{3D Hand Estimation Module}
We present a monocular 3D hand pose estimation module, denoted by $M_H$, estimating the parameters of the hand model $H$. 
In particular, our hand module is inspired by the recently proposed monocular body pose estimation approaches~\cite{kanazawa2018end, humanMotionKanazawa19, kolotouros2019spin}, thus follows the similar model architecture, parameterizations, and training stages.
Leveraging the achievement in body pose estimation area, we found that our hand pose estimation method can be robustly applicable for various in-the-wild situations, showing the state-of-the-art performance in public hand pose estimation benchmarks.

\noindent \textbf {Hand Module Architecture.} 
Our hand module $M_H$ is built upon an end-to-end deep neural network architecture to regress the hand pose parameters defined in Eq.~\eqref{eq:smplx_hand}. Our hand module $M_H$ is defined as:
\begin{equation}
\label{eq:hand_module}
[\boldsymbol{\phi}_h, \boldsymbol{\theta}_h, \boldsymbol{\beta}_h, \boldsymbol{c}_h]  = M_H(\mathbf{I}_H), \\
\end{equation}
where $\mathbf{I}_H$ is an input RGB image cropped around a hand region. $\boldsymbol{c}_h = (\boldsymbol{t}_h, s_h)$ is weak-perspective camera parameters which allows to project a posed 3D hand model to an input image.
Here, $\boldsymbol{t}_h \in  \mathbb{R}^{2}$ is for for 2D translation on the image plane, and $s_h \in  \mathbb{R}$ is a scale factor. Thus, the $i$-th 3D hand joint,$\boldsymbol{J}^{3D}_{h,i}$ can be projected as:
\begin{equation}
\boldsymbol{J}^{2D}_{h,i}= \boldsymbol{s}_h \Pi (\boldsymbol{J}^{3D}_{h,i}) +\boldsymbol{t}_h,
\end{equation}
where $\Pi$ is an orthographic projection.


%
Following the body pose estimation approaches~\cite{kanazawa2018end, kolotouros2019spin}, the architecture of our hand module $M_H$ is composed of an encoder and a decoder structure, where the encoder outputs the encoded features from input images, and the decoder regresses the hand pose parameters from the features. See Figure~\ref{fig:hand_model} for the overview of our hand module. 
We use the ResNet-50~\cite{He:2016:Deep} for the encoder network. The decoder network is composed of a group of fully connected layers.
Our hand module is trained with the data for the right hand. The images and annotations for the left hand are used after vertical flipping. During the testing time, the left hand images are flipped and processed as if they were a right hand, and their outputs are flipped back to the original left hand space.

Note that the shape parameter $\boldsymbol{\beta}_h$ is originally defined for whole body model $\boldsymbol{\beta}_w$, but we only consider the deformation for the hand vertices defined in~\ref{eq:smplx_hand}, ignoring the body part. 
We describe how this can be handled in our integration module.

\begin{figure}[t]
	\begin{center}
		\includegraphics[width=0.8\linewidth]{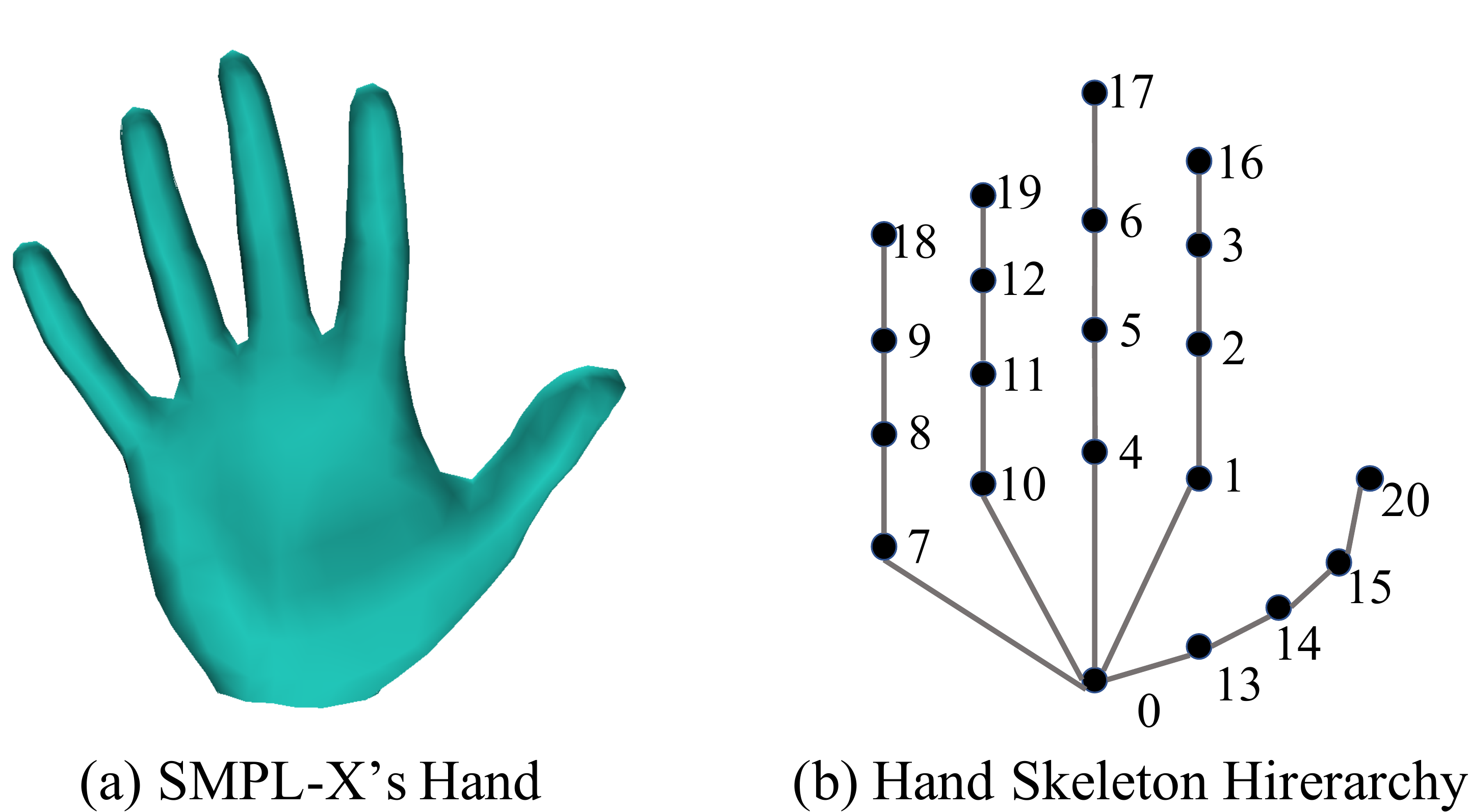}
	\end{center}
	\vskip -0.3cm
	\caption{\small Our hand model taken from SMPL-X. We take the hand part of SMPL-X as a stand-alone hand model for hand pose estimation. The example mesh is shown in (a) and the skeleton hierarchy is shown in (b).}
	\label{fig:hand_visualize}
	\vspace{0.1cm}
\end{figure}

\begin{figure}[t]
	\begin{center}
		\includegraphics[width=1.0\linewidth]{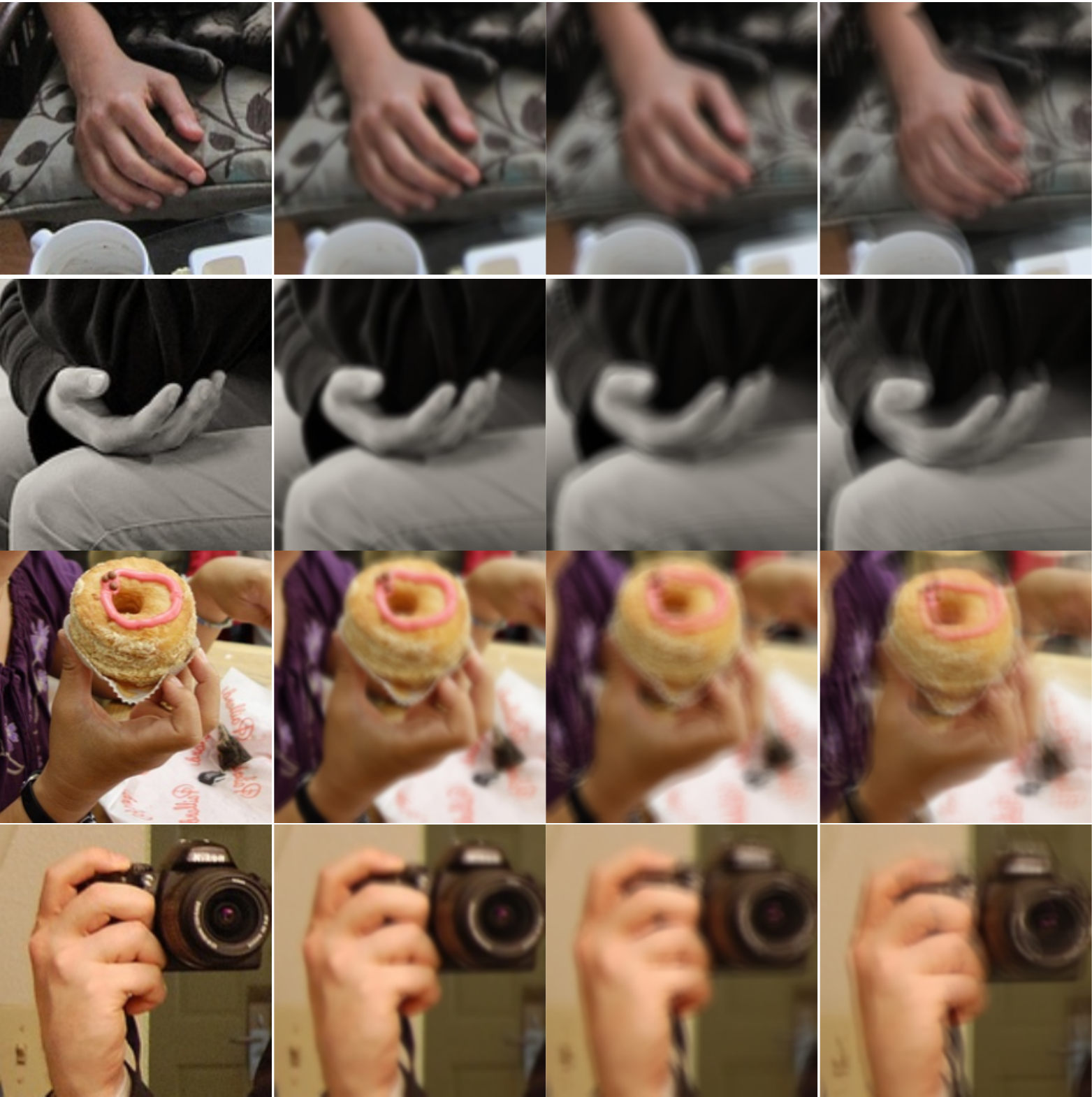}
	\end{center}
	\vskip -0.3cm
	\caption{\small Motion Blur Augmentation. We show example images of motion blur augmentation. From left to right: original images, augmented images after applying different motion blur kernels.}
	\label{fig:motion_blur}
	\vspace{0.1cm}
\end{figure}

\noindent \textbf{Training Method.} 

We consider three different types of annotations: (1) 3D pose annotations (in angle-axis representation), (2) 3D keypoint (joint) annotations, and (3) 2D keypoint annotations. 
The losses for each of the annotations, namely $L_{\boldsymbol{\theta}}$, $L_{3D}$ and $L_{2D}$, are defined as follows:

\begin{equation}
\label{eq:3d_2d_loss}
\begin{aligned}
& L_{\theta} = \lVert \boldsymbol{\theta}_h - \hat{\boldsymbol{\theta}_h} \rVert^{2}_{2}, \\
& L_{3D} =  \lVert \boldsymbol{J}^{3D}_{h} - \boldsymbol{\hat{J}}^{3D}_{h} \rVert^{2}_{2}, \\
& L_{2D} = \lVert \boldsymbol{J}^{2D}_{h} - \hat{\boldsymbol{J}}^{2D}_{h}) \rVert, \\
\end{aligned}
\end{equation}
where $\hat{\boldsymbol{\theta}_h}$, $\hat{\boldsymbol{J}}^{3D}_{h}$ and  $\hat{\boldsymbol{J}^{2D}_{h}}$ are the ground-truth annotations of angle-axis pose parameters, 3D keypoints, and 2D keypoints.
In particular, the 2D keypoint loss $L_{2D}$ is necessary to estimate the camera projection parameters.
We do not use the shape parameters provided by the 3D hand datasets such as FreiHAND~\cite{Zimmermann:2019:FreiHAND}, since these are defined for the MANO model~\cite{romero2017embodied} and not compatible with our hand model from SMPL-X.
Instead, an additional shape parameter regularization loss $L_{reg}$ is applied:
\begin{equation}
\label{eq:reg_loss}
\begin{aligned}
& L_{reg} = \lVert \boldsymbol{\beta}_h \rVert^{2}_{2}.\\
\end{aligned}
\end{equation}

The overall loss $L$ used to train our hand module is defined as follows:
\begin{equation}
\label{eq:overall_loss}
\begin{aligned}
	& L = \lambda_{1}L_{\boldsymbol{\theta}} + \lambda_{2}L_{3D} + \lambda_{3}L_{2D} + \lambda_{4}L_{reg}.
\end{aligned}
\end{equation}
In experiments, the balanced weights are set as $\lambda_{1} = 10$, $\lambda_{2} = 100$, $\lambda_{3} = 10$ and $\lambda_{4} = 0.1$.


\noindent \textbf{Datasets Preprocessing.}
3D hand pose datasets are often built by multi-view setups in controlled environments to obtain ground-truth annotations. 
A model trained with these datasets often suffer from overfitting, showing limited performance when applied to outdoor in-the-wild data. 
Notably, recent 3D body pose estimation approaches have shown that leveraging diverse datasets can greatly improve its generalization ability~\cite{kanazawa2018end, humanMotionKanazawa19, rong2019delving, kolotouros2019spin}. 
Following this, we include as many publicly available datasets as possible towards in-the-wild 3D hand pose estimation. 
More details of the datasets are discussed in section~\ref{sec:hand_datasets}.
The major challenge in using diverse datasets is that their annotation types vary. For example, there exist available ground-truth joint angle parameters in FreiHand~\cite{Zimmermann:2019:FreiHAND} and HO-3D~\cite{Hampali:2019:HO3D} datasets, while others do not contain it. 
Furthermore, the details of hand annotations including skeleton hierarchy and scales are also different across datasets. 
To handle this, we perform several pre-processing steps to make them consistent and compatible with our hand model, including 
1. Rescaling all 3D keypoint annotations to be compatible with our hand model, by using the middle finger's knuckle length\footnote{the skeleton between $4$-th and $5$-th joints efined in Figure~\ref{fig:hand_visualize}.} as a reference. 
2. Re-ordering the 3D keypoints joints to be the same as our hand model's skeleton hierarchy shown in Figure~\ref{fig:hand_visualize}.
\noindent \textbf{Training Data Augmentation.} 
Performing data augmentations during training is a common practice to enable model with better generalization ability. Following previous approaches~\cite{Zhang:2019:End}, we apply common data augmentation strategies including random scale, random translation, color jittering, and random rotation. 

Importantly, we recognize that in-the-wild videos are often accompanied by severe motion blur. To achieve robustness to motion blur, we additional apply motion blur augmentation to the images. We first use the methods in previous papers~\cite{boracchi2010uniform,boracchi2012modeling} to generate blur kernels and then use 2D filtering to add blurriness to images. 
The experiments show that our motion blur augmentation is beneficial to generalize our hand module for in-the-wild scenes.
Examples of motion blur augmentation are shown in Figure~\ref{fig:motion_blur}.

\subsection{3D Body Estimation Module}
We leverage the state-of-the-art monocular 3D pose estimation methods~\cite{kanazawa2018end, kolotouros2019spin} with a few modifications for our body estimation module. 
The recent monocular body pose estimation methods~\cite{kanazawa2018end,kolotouros2019spin} are based on the SMPL~\cite{Loper2015} model to capture torso and limb motion. 
Although our body module produces similar outputs, these previous methods cannot be directly applicable for our objective, since SMPL model's shape parameters are not compatible with SMPL-X.  Thus, we fine-tune the publicly available state-of-the-art pose estimator~\cite{kolotouros2019spin} by replacing SMPL with SMPL-X in the training pipelines. 
For training, we use the publicly available indoor 3D pose datasets such as human3.6M~\cite{h36m_pami}.
We also include the pseudo-ground truth annotations introduced in \cite{Joo:2020:Exemplar} that provides SMPL fitting paired with the in-the-wild 2D keypoint datasets (\eg, COCO~\cite{lin2014microsoft} and MPII~\cite{Andriluka-14}). 
Since all these annotations are in SMPL format, we ignore the shape parameters of ground truth SMPL annotations, and only use the pose parameters and 2D keypoint annotations that are compatible to SMPL-X model. 
We follow the same neural network architecture with the similar training steps as in the work of \cite{kolotouros19convolutional}, without using the SMPLify part.

Our body module $M_B$ produces the torso and limb parameters defined in Eq.~\ref{eq:smplx} from an single image:

\begin{equation}
\label{eq:body_inference}
[\boldsymbol{\phi}_b, \boldsymbol{\theta}_b, \boldsymbol{\beta}_b, \boldsymbol{c}_b]  = M_B(\boldsymbol{I}_b), \\
\end{equation}
where $\boldsymbol{I}_b$ is an input image cropped around a target single person's whole body. 
Similar to Eq.~\ref{eq:smplx}, $\boldsymbol{\phi}_b \in \mathbb{R}^{3}$ is the global body orientation, $\boldsymbol{\theta}_b  \in \mathbb{R}^{21 \times 3}$ is the body pose parameters (without any hand joints), and $\boldsymbol{\beta}_b \in \mathbb{R}^{10}$ is the shape parameter. Again, $\boldsymbol{\beta}_b$ shares the same parameterization space as $\boldsymbol{\beta}_w$, which is defined in Eq.~\ref{eq:smplx}.
Similar to Eq.~\ref{eq:hand_module}, we use weak perspective camera parameters  $\boldsymbol{c}_b = (\boldsymbol{t}_b, \boldsymbol{s}_b)$.

Note that the existing body pose estimators including our fine-tuned version do not accurately estimate the the wrist and arm orientation due to inaccurate or insufficient annotations (\eg, only one keypoint is annotated for a wrist), as shown in Figure~\ref{figure:wholebody_optimization}. Our integration module solves this issue. 

\subsection{Whole Body Integration Module}
Our integration module combines the outputs from the 3D body and hand modules into a unified representation as a form of SMPL-X model. For the integration, we present two strategies: (1) a fast method by simple copy-and-paste composition, and (2) an optimization framework to include additional 2D keypoint cues for more accurate output.

\noindent \textbf{Fast Body and Hand Composition by Copy-and-Paste.}
%
Since the outputs from our hand and body modules are compatible with the SMPL-X model, they can be easily combined as the single form. A simple strategy is just transferring the corresponding joint angle parameters from the outputs of each of the hand and body modules. However, the wrist parts require additional processing, because we obtain two different outputs from the body and hand modules (represented by the global hand orientation $\boldsymbol{\phi}_h$). Let us denote the pose parameters for the wrist joint as $\boldsymbol{\theta}^{\text{wrist}}$,  then $\boldsymbol{\theta}_b = \boldsymbol{\tilde{\theta}}_b \cup \{  \boldsymbol{\theta}_b^{\text{rwrist}}, \boldsymbol{\theta}_b^{\text{lwrist}} \} $, where $\boldsymbol{\tilde{\theta}}_b$ includes all body pose parameters except wrists. We use the similar notations for the whole body pose parameters, $\boldsymbol{\theta}_w^{\text{rwrist}}$, $\boldsymbol{\theta}_w^{\text{lwrist}}$, and $\boldsymbol{\tilde{\theta}}^b_w$. Then, whole body integration by copy-and-paste can be performed as:

\begin{equation}
\label{eq:copy_paste}
\begin{aligned}
& \boldsymbol{\phi}_w = \boldsymbol{\phi}_b, \\
& \boldsymbol{\beta}_w = \boldsymbol{\beta}_b,\\
& \boldsymbol{c}_w = \boldsymbol{c}_b,
\end{aligned}
\end{equation}


\begin{equation}
\label{eq:copy_paste2} 
\begin{aligned}
\left( \boldsymbol{\tilde{\theta}}_w^b, \boldsymbol{\theta}_w^{lh}, \boldsymbol{\theta}_w^{rh} \right) & = \left( \boldsymbol{\tilde{\theta}}_b, \boldsymbol{\theta}_{lh}, \boldsymbol{\theta}_{rh} \right), \\
\left( \boldsymbol{\theta}_w^{\text{lwrist}}, \boldsymbol{\theta}_w^{\text{rwrist}} \right) & = \left( \Gamma_l \left( \boldsymbol{\theta}_b, \boldsymbol{\phi}_{lh}   \right), \Gamma_r \left( \boldsymbol{\theta}_b, \boldsymbol{\phi}_{rh}   \right)  \right), \\
\end{aligned}
\end{equation}

where $\Gamma_l$ and $\Gamma_r$ are the functions to convert the global wrist orientation $\boldsymbol{\phi}_{h}$ obtained from the hand module to the local wrist pose parameters w.r.t. its parent joint in the SMPL-X skeleton hierarchy. This can be implemented by comparing $\boldsymbol{\phi}_{h}$ with the global orientation of the current wrist pose from $\boldsymbol{\theta}_b$ that can be computed by following the forward kinematics of the body skeleton hierarchy.
This strategy requires almost no extra computation, making our separate modules to contribute a common whole body model simultaneously. We found this simple integration produces convincing results, especially for the scenarios with computational bottlenecks as in our live demo. 

\begin{figure}
  \centering
  \includegraphics[width=\linewidth]{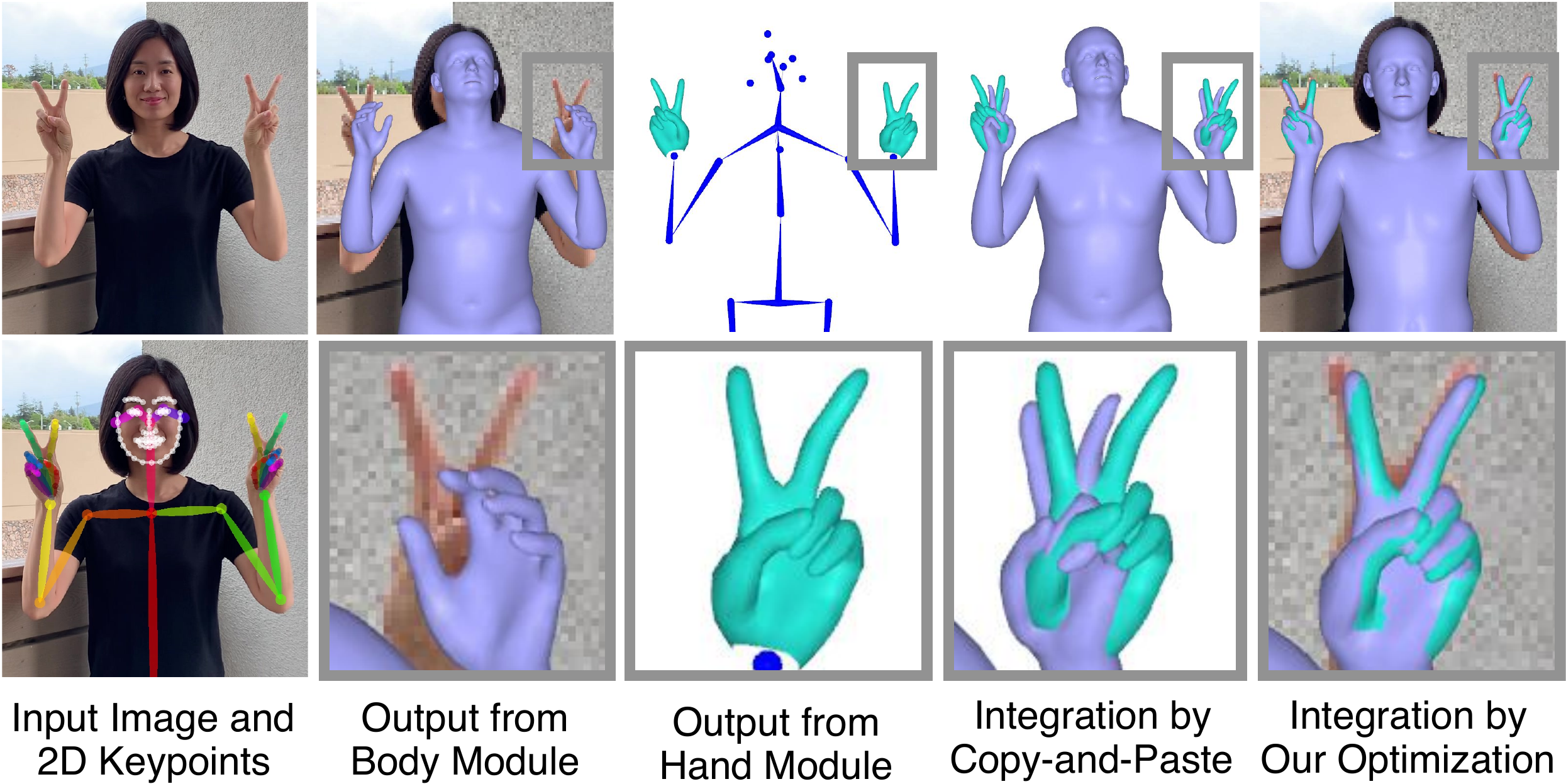}
  \caption{\small Optimizing the whole body model (SMPL-X) with 3D hand prediction and 2D keypoint estimation. (a) An input image and the estimated 2D keypoints by OpenPose~\cite{Cao:2019:Openpose}; (b) 3D body pose estimation from our body module; (c) The output of 3D hand module aligned to the wrist joints of SMPL-X; (d) Integration output by copy-and-paste strategy; (e) Integration output by our optimization framework.}
  \label{figure:wholebody_optimization}
\end{figure}

\noindent \textbf{Hand and Body Composition via Optimization.}
%
As an alternative integration method, we build an optimization framework to fit the whole body model parameters given the outputs from body and hand modules. This strategy is particularly helpful to reduce the artifact around the wrist parts over the copy-and-paste strategy, and also can take advantage from the 2D keypoint estimation output~\cite{Cao:2019:Openpose} for better 2D localization quality. In particular, our optimization framework finds the whole body model parameters that minimize the following objective cost function:
\begin{equation}
    \begin{gathered}
        \mathcal{F}([\boldsymbol{\phi}_w, \boldsymbol{\theta}_w,   \boldsymbol{\beta}_w, \boldsymbol{c}_w]) = \mathcal{F}^{2d} + \mathcal{F}^{pri},
        \label{eq:full}
    \end{gathered}
\end{equation}
where $\mathcal{F}^{2d}$ is the 2d reprojection cost term between the 2D keypoint estimation~\cite{Cao:2019:Openpose} and the projection of 3D joints (body and both hands), and the prior term $\mathcal{F}^{pri}$ is needed to keep the 3D pose and shape parameters in plausible space, as in SMPLify method\cite{Bogo2016}. 
We first initialize all parameters by our copy-and-paste strategy except that we do not apply $\Gamma$ to transfer the global hand orientation to whole body model. Instead, the wrist orientations of the hands can be obtained by minimizing the Eq.~\eqref{eq:full} with other parameters. 
While a Gaussian mixture model learnt from motion capture dataset~\cite{CMUMocap} is often used for the body pose prior term as in SMPLify method~\cite{Bogo2016}, we use the the exemplar fine-tuning approach introduced in \cite{Joo:2020:Exemplar} for the similar goal, by applying neural network fine-tuning of $M_B$ for each frame independently, which does not require additional regularization term but still keep the 3D pose in plausible space. 
Note that our optimization framework requires only a few iteration (20 iterations in all our experiments), since outputs from the body and hand modules output is already close to the target status. See Figure~\ref{figure:wholebody_optimization} for the example of our optimization.

\section{Experiments} \label{sec:Results}

In this section, we first describe the implementation details. Then we summarize the datasets used for our hand module training. After that, we quantitatively and qualitatively compare our methods with the state-of-the-art approaches. We also perform ablation studies to examine the key designs of our methods. 

\subsection{Implementation Details}
\label{sec:implementation_detail}

\noindent \textbf{Bounding Boxes.}
For the online version, we use OpenPose~\cite{Cao:2019:Openpose} to obtain body bounding boxes. After processing the body, the hand bounding boxes are obtained by projecting the hand part of the estimated 3D body to image space.
For the offline processing of internet videos, we use OpenPose detections to localize both bodies and hands.

\noindent \textbf{Video Processing.}
For copy-and-paste strategy (used in online demo and offline internet videos), the videos are processed frame-by-frame without any post-processing.
For optimization-based strategy, after obtaining per-frame outputs, we apply a naive temporal smoothing for each separate dimension of parameters (shape, pose, and
camera). We use a 5-frame-size smoothing kernel with the weight $\left [0.1, 0.2, 0.5, 0.2,
0.1\right ]$.
It is noted that our copy-and-paste method can generate temporally-stable results even without smoothness. We believe it is due to the fact that recent CNN pose regressors tend to produce
such output as demonstrated in recent papers (\eg SPIN~\cite{kolotouros19convolutional}), thanks to multiple augmentation tricks in training. 
The optimization-based method (SMPLify-X and MTC) suffers from temporal instability due to the complicated optimization procedures with multiple-stages (\eg torso first and others later) and elaborated balancing issues between data term and prior term. 
The processing time of each method are compared in Table~\ref{tab:runtime}. The processing time of our copy and paste method is about 9.5 fps, where the code is implemented in python and runs in a single GeForce RTX 2080 GPU. In our supplementary video, we also show a live demo using a single webcam, which cannot be performed by  alternative approaches.

\begin{table}[t] \centering \small %
	\caption{\small Processing time of various methods.}
	\setlength\tabcolsep{2pt}
	\vskip -0.2cm
	\begin{tabular}{c|c|c|c|c|c}
		\hline
		Method $\rightarrow$         & \multirow{2}{*}{SMPLify-X} & \multirow{2}{*}{MTC}    &  Online   &  Offline   & Offline  \\
		Time (fps)	$\downarrow$	 &							   &                        &    (CP) 	&  (CP)      & (OP)     \\
		\hline
		Preprocess (fps)     		 & 7.5       & 7.5    &  35        & 7.5 			  &      7.5        \\
		Model 		(fps)     		 & 0.01      & 0.1    &  13        & 13              &      1.1      \\
		Overal      (fps)    		 &  0.01      & 0.1    & 9.5        & 4.7			  &     0.95      \\
		\hline
	\end{tabular}	
	\label{tab:runtime}
\end{table}

\noindent \textbf{Hand Module.}
Input images of the hand module are center-cropped surrounding the hands, where the bounding boxes for cropping are given by 2D hand keypoints. Ground-truth 2D keypoints are used for training time, the predicted keypoints from OpenPose~\cite{Cao:2019:Openpose} are used for testing time. 
The cropped images are further padded and resized to size of $224 \times 224$.
During training, we apply data augmentations to each of training images via random scaling, translation, rotation, color jittering, and synthetic motion blur.
The hand module architecture is based on ResNet-50~\cite{He:2016:Deep} with two additional fully connected layers to map the output features of ResNet to vectors with $61$ dimension, which is composed of camera parameters $C$ (3 dimensions), hand global rotation $\phi_h$ (3 dimensions), hand pose parameters $\theta_{h}$ (45 dimensions) and shape parameters $\beta_{h}$ (10 dimensions). 
The hand module is implemented with PyTorch~\cite{Paszke:2019:Pytorch}. The Adam optimizer~\cite{Kingma:2016:Adam} with learning rate $1\mathrm{e}{-4}$ is used to train the model. The hand module is trained until converge, which takes about $100$ epochs.

\noindent \textbf{Body Module.}
We follow the similar training steps to the state-of-the-art method~\cite{kolotouros2019spin} using the Human3.6M~\cite{h36m_pami} and COCO datasets~\cite{lin2014microsoft} with the pseudo SMPL annotations~\cite{Joo:2020:Exemplar}. 
Our training starts from the pre-trained model of SPIN~\cite{kolotouros2019spin} with substituting the SMPL parameters to the SMPL-X parameters. 
The model is then finetuned using the Adam optimizer~\cite{Kingma:2016:Adam} with learning rate $5\mathrm{e}{-5}$ for about $20$ epochs.
Besides, we use neutral SMPL-X model for both hand and body module.

\subsection{Datasets}
\label {sec:hand_datasets}

\noindent \textbf{FreiHAND}.
FreiHAND~\cite{Zimmermann:2019:FreiHAND} is a dataset with ground truth 3D hand joints and MANO parameters for real human hand images. 
The 3D annotations are obtained by a multi-camera system and a semi-automated approach. The obtained data is further augmented with synthetic backgrounds.
In our experiments, we randomly select 80\% of samples from original training set as training data and use the remaining 20\% of samples for validation.

\noindent \textbf{HO-3D}. 
HO-3D dataset~\cite{Hampali:2019:HO3D} is a dataset aiming to study the interaction between hands and objects. 
The dataset has 3D joints and MANO pose parameters for hands, and also has 3D bounding boxes for objects the hands interact with. In this paper, we only use 3D annotations of hands.
The training set is composed of different sequences, each of which records one type of hand-object interaction.
Following the similar practice in processing FreiHAND, we randomly choose 80\% of sequences from the original training set as training data and use the remaining 20\% of sequences for validation.

\noindent \textbf{MTC.} 
Monocular Total Capture~\cite{Xiang:2019:Monocular} is a dataset captured by Panoptic Studio~\cite{joo2015panoptic,joo2017panoptic} in a multi-view setup with 30 HD cameras. It has 3D hand joints annotations for both body and hands. The sequences are mainly the range of motion data of multiple subjects.
To polish the dataset, we filter out erroneous samples where hands are not visible or too small. 

\noindent \textbf{STB.} 
Stereo Hand Pose Tracking Benchmark~\cite{Zhang:2017:Stereo} is composed of 15,000 training samples and 3,000 testing samples. The provided annotations include 3D joints and depth images. 
In our experiments, we use 3D joints only.
We use training set of STB to train our model and compare with other state-of-the-art methods on the validation set.
To unify definition of joints, following the practice of~\cite{Cai_2018_ECCV,Ge:2019:3D}, we move the root joint from palm center to wrist.

\noindent \textbf{RHD.} 
Rendered Hand Dataset~\cite{Zimmermann:2017:Learning} is a synthetic dataset that has 2D and 3D hand joint annotations. It is composed of 41,258 training samples and 2,728 testing samples. We train our model on the training set and compare with other state-of-the-art methods on the testing set.

\noindent \textbf{MPII+NZSL.} 
MPII+NZSL dataset~\cite{simon2017hand} is composed of in-the-wild images with manually annotated 2D hand joints. 
It includes challenging images with occlusion, blur, and low resolution. To show our models' generalization ability, our models are no trained on the MPII+NZSL, we only use it for validation.

\begin{figure*}[t]
	\begin{center}
		\includegraphics[width=0.95\linewidth]{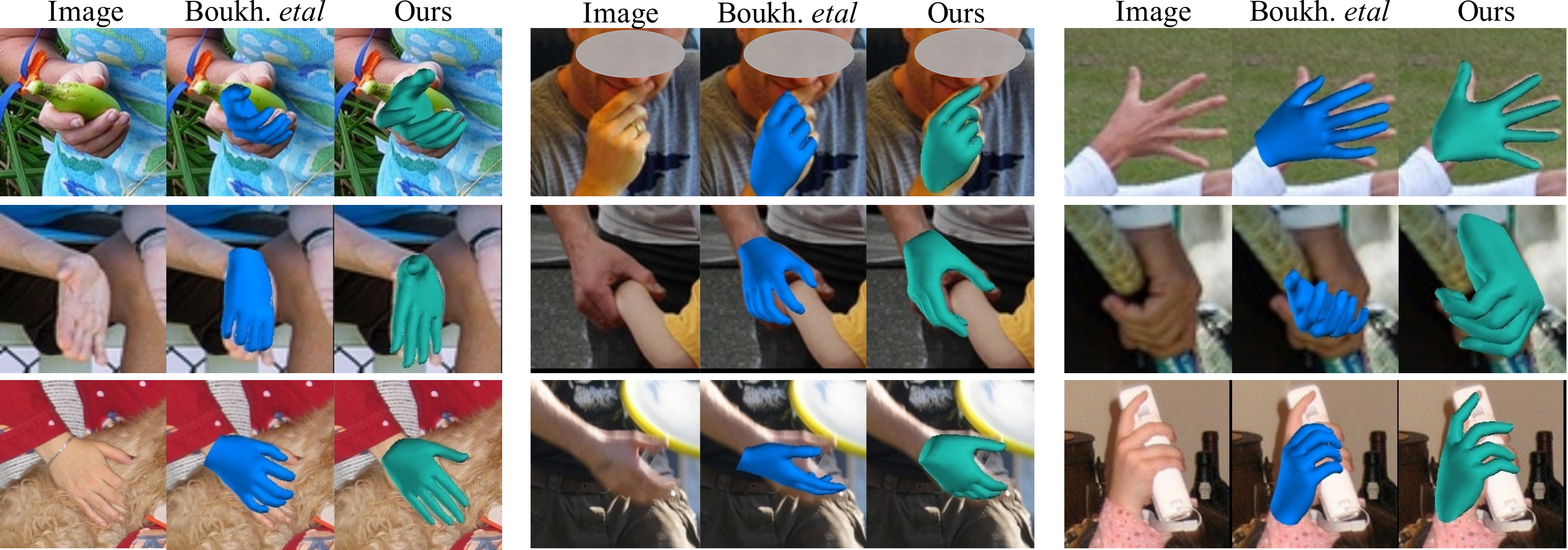}
	\end{center}
	\vskip -0.5cm
	\caption{\small Qualitative comparison with State-of-the-are methods. The images are selected from COCO dataset~\cite{lin2014microsoft}. We qualitatively compare our models performance with Boukhayma~\etal~\cite{Boukhayma:2019:3D}.}
	\label{fig:compare_sota_hand}
	\vspace{-0.1cm}
\end{figure*}

\begin{figure*}[t]
	\begin{center}
		\includegraphics[width=0.95\linewidth]{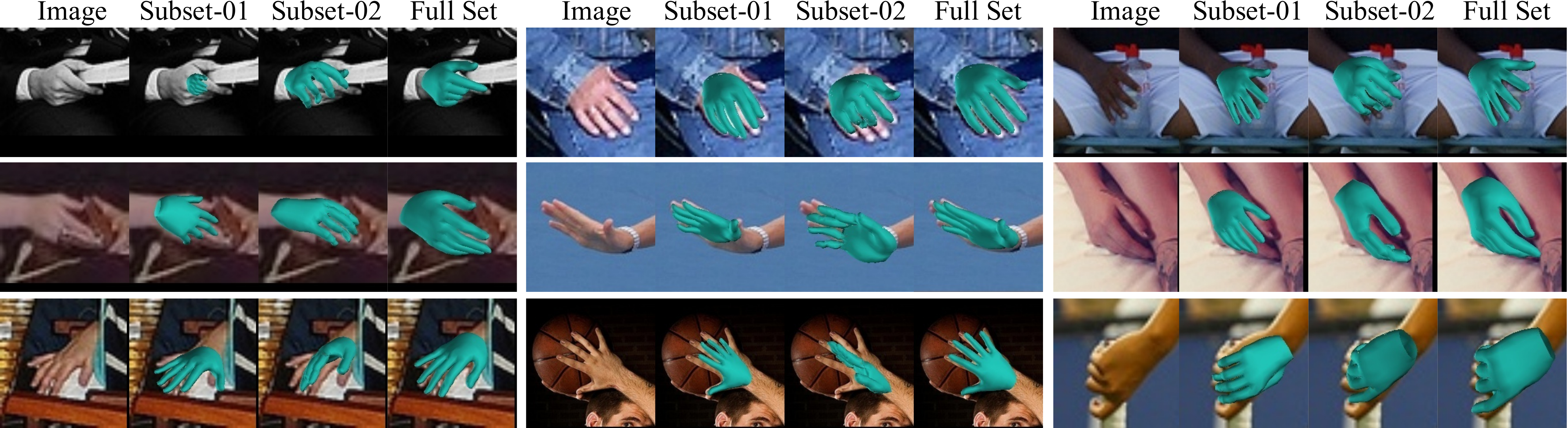}
	\end{center}
	\vskip -0.5cm
	\caption{\small Ablation study on training dataset. We show qualitative ablation study on using different datasets in training our hand model. ``Subset-01'' means using the combination of datasets FreiHADN~\cite{Zimmermann:2019:FreiHAND} and HO-3D~\cite{Hampali:2019:HO3D}. ``Subset-02'' means using the combination of datasets: STB~\cite{Zhang:2017:Stereo}, RHD~\cite{Zimmermann:2017:Learning} and MTC~\cite{Xiang:2019:Monocular}. ``Full set'' means using all the above datasets. The images are selected from COCO dataset~\cite{lin2014microsoft}.}  
	\label{fig:ablation_dataset}
	\vspace{-0.1cm}
\end{figure*}

\begin{figure*}[t]
	\begin{center}
		\includegraphics[width=0.95\linewidth]{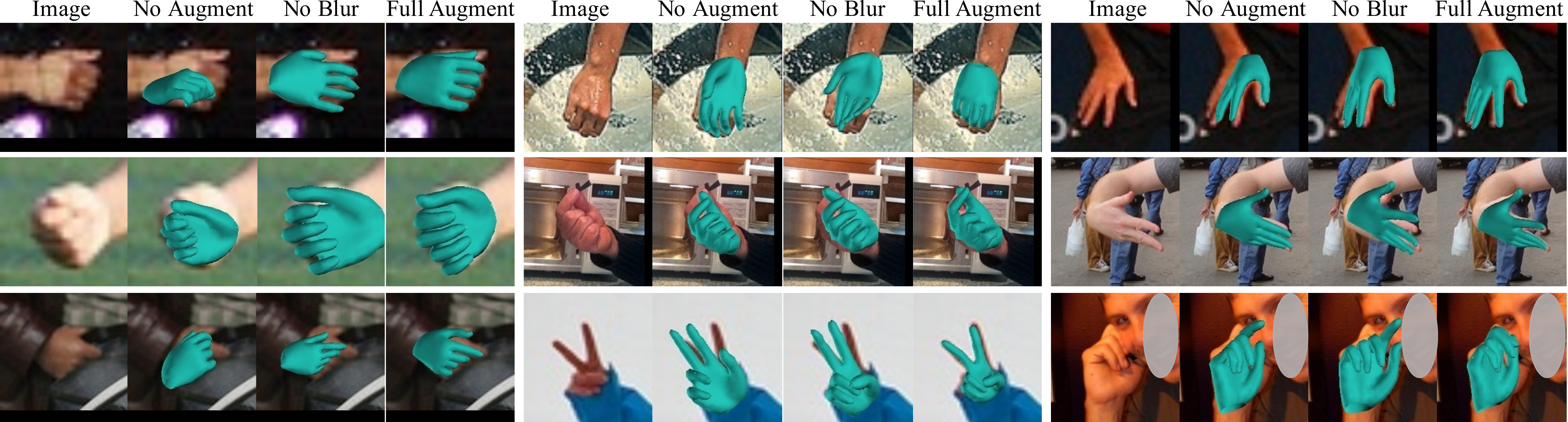}
	\end{center}
	\vskip -0.5cm
	\caption{\small Ablation study on data augmentation. We show qualitative ablation study on training our hand model using different data augmentation. ``No Blur'' refers to model trained with all data augmentation strategies except motion blur augmentation. ``Full Augment'' refers to model trained with all data augmentation strategies. The images are selected from COCO dataset~\cite{lin2014microsoft}.}
	\label{fig:ablation_augment}
	\vspace{-0.1cm}
\end{figure*}



\begin{figure*}[t]
	\begin{center}
		\includegraphics[width=0.95\linewidth]{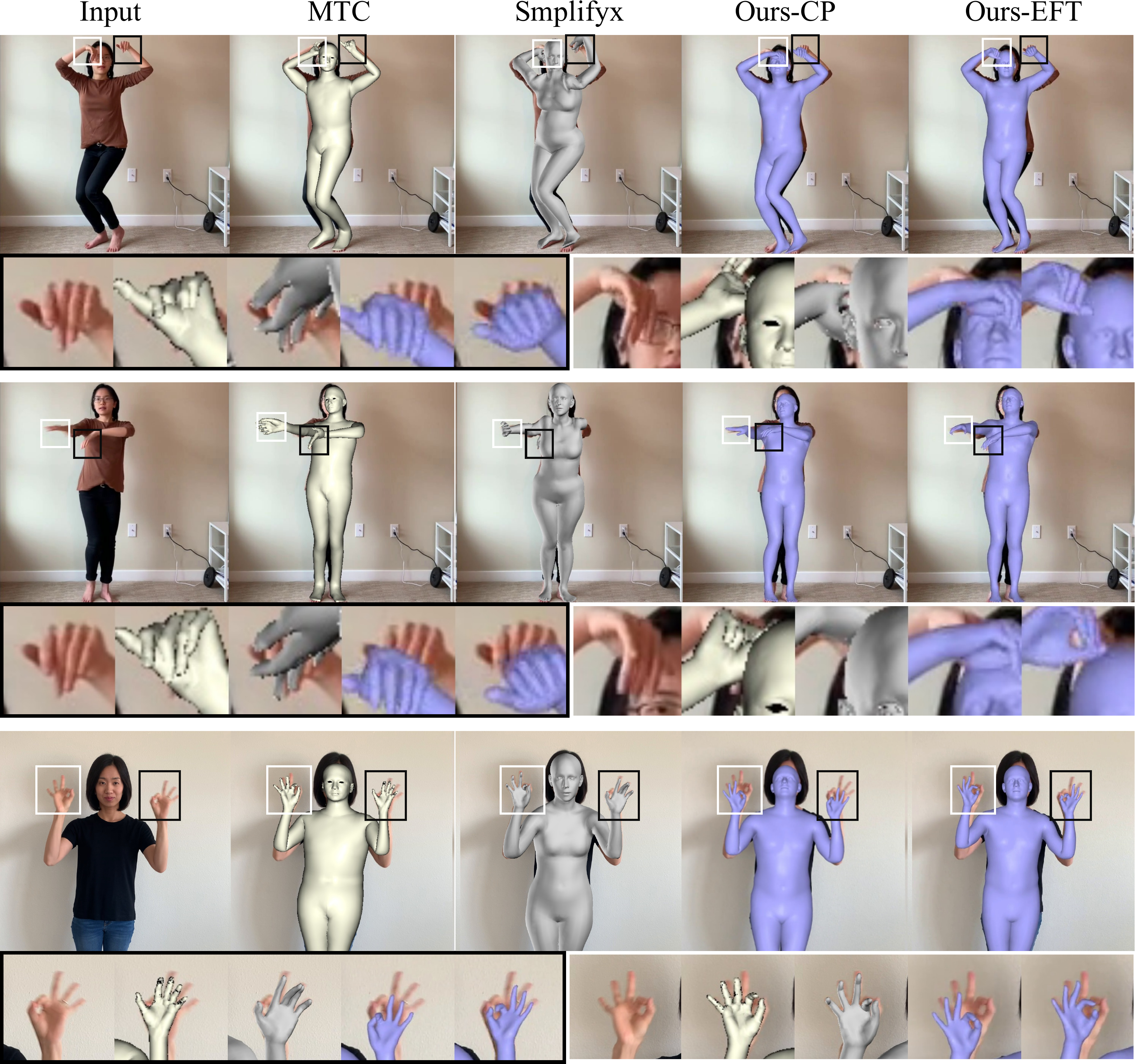}
	\end{center}
	\vskip -0.3cm
	\caption{\small Qualitative comparison over the previous whole body pose estimation methods. We compare our method with MTC~\cite{Xiang:2019:Monocular} and SMPLify-X~\cite{Pavlakos:2019:SMPLX}. Our method from two different strategies, by the copy-and-paste method (Ours-CP) and by an optimization framework (Ours-OP), is not only faster, but also produces more convincing whole body motion capture outputs.}
	\label{fig:compare_sota_body}
	\vspace{0.3cm}
\end{figure*}

%
%

\subsection{Hand Module Evaluation}

\begin{table}[t] \centering \small %
	\caption{\small Comparison of our hand module with the state-of-the-art hand methods on three public benchmarks, namely STB, RHD and MPII+NZSL. For STB and RHD, we use 3D AUC as the evaluation metric. The threshold ranges from 20mm to 50mm. For MPII+NZSL, we use 2D AUC as the evaluation metric. The threshold ranges from 0px to 30px.}
	\vskip -0.3 cm
	\setlength\tabcolsep{2pt}
	\begin{tabular}{c|c|c|c}
		\hline
		Dataset $\rightarrow$    & \multirow{2}{*}{STB}  & \multirow{2}{*}{RHD} & \multirow{2}{*}{MPII+NZSL} \\
		Method $\downarrow$		 & 	 		&         &   \\   	
		\hline
		\cite{Zimmermann:2017:Learning}	    & 0.948	 & 0.675   & 0.171 \\
		\cite{Boukhayma:2019:3D}   				& 0.994  &  -      & 0.501 \\
		\cite{baek2019pushing}                  &   -    & 0.926   &   -   \\
		\cite{Ge:2019:3D}         						& \textbf{0.995}  & 0.92    & 0.15  \\
		\cite{Xiang:2019:Monocular}       		    & 0.994  & -       & 0.340 \\
		\cite{Zhang:2019:End}                       & \textbf{0.995}  & 0.901   & -     \\
		\hline
		Ours-less-datasets                                      & 0.992  & 0.918   & 0.556  \\
		Ours-no-data-augment                                    & 0.991  & 0.893   & 0.608   \\
		Ours-no-shape-params                                    & 0.987  & 0.910   & 0.647   \\ 
		Ours            							 			& 0.992  & \textbf{0.934}   & \textbf{0.655}  \\
		\hline
	\end{tabular}	
	\label{tab:hand_compare_with_sota}
\end{table}

\begin{table}[t] \centering \small %
	\caption{\small Ablation study on dataset. We show the results of our hand module trained with different datasets. These models are evaluated on MPII+NZSL~\cite{simon2017hand} using 2D AUC as metric. For data augmentation, we use all the available datasets.}
	\setlength\tabcolsep{2pt}
	\vskip -0.2cm
	\begin{tabular}{c|c||c|c|c||c}
		\hline
		FreiHAND     & HO-3D     & MTC      &   STB   & RHD     & MPII+NZSL \\
		\hline
		\cmark       &           &          &         &         & 0.482 \\
		             & \cmark    &          &         &         & 0.367 \\
		\cmark       & \cmark    &          &         &         & 0.526 \\
		             &           & \cmark   & \cmark  & \cmark  & 0.556 \\
		\cmark       & \cmark    & \cmark   &         &         & 0.595 \\
		\cmark       & \cmark    & \cmark   & \cmark  &         & 0.598 \\
		\cmark       & \cmark    & \cmark   &         & \cmark  & 0.645 \\
		\cmark       & \cmark    & \cmark   & \cmark  & \cmark  & 0.655 \\
		\hline
	\end{tabular}	
	\label{tab:ablation_dataset}
\end{table}

\begin{table}[t] \centering \small %
	\caption{\small Ablation study on data augmentation. We show the results of our hand module trained with different data augmentation strategies. These models are evaluated on MPII+NZSL~\cite{simon2017hand} using 2D AUC as metric.} 
	\vskip -0.3cm
	\begin{tabular}{c|c|c|c|c||c}
		\hline
		\multirow{2}{*}{Position} & \multirow{2}{*}{Rescale} & Color & \multirow{2}{*}{Rotation} & Motion & MPII + \\
							      &							& Jittering & 						&	Blur  & NZSL \\
		\hline    
		\xmark       & \xmark    & \xmark             & \xmark       & \xmark      & 0.608            \\
		\cmark       & \cmark    &                    &              &             & 0.610 \\
		\cmark       & \cmark    & \cmark             &              &             & 0.618 \\
		\cmark       & \cmark    & \cmark             & \cmark       &             & 0.622 \\
		\cmark       & \cmark    & \cmark             & \cmark       & \cmark      & 0.655 \\
		\hline
	\end{tabular}	
	\vspace{0.3cm}
	\label{tab:ablation_data_augmentation}
\end{table}


\noindent \textbf{Comparison with State-of-the-art Methods.}
We compare our hand module with the previous state-of-the-art hand approaches on three public hand benchmarks, STB~\cite{Zhang:2017:Stereo}, RHD~\cite{Zimmermann:2017:Learning} and MPII+NZSL~\cite{simon2017hand}. 
For each validation dataset, we calculate the percentage of correct keypoints (PCK) under different thresholds and calculate the corresponding Area Under Curve (AUC) for PCK. 
For STB~\cite{Zhang:2017:Stereo} and RHD~\cite{Zimmermann:2017:Learning}, we use 3D AUC and the threshold ranges from 20mm to 50mm. 
For MPII+NZSL~\cite{simon2017hand}, we use 2D AUC and the threshold ranges from 0px to 30px.

The results are listed in Table~\ref{tab:hand_compare_with_sota}. For fair comparison, all the methods takes single RGB image as input.
``Ours'' refers to our best model trained with all the datasets and all data augmentation srategies. 
It outperforms previous methods on RHD and MPII+NZSL and shows a comparable performance in STB. 
Notably, our method shows significantly better 2D localization accuracy on challenging in-the-wild dataset MPII+NZSL, demonstrating its generalization ability to in-the-wild scenarios. 

%


We also compare our own best model with variants of our method. 
%
``Ours-no-shape-params'' differs from ``Ours'' in that the shape parameters $\beta$ are not used and fixed to zero. 
``Ours-no-data-augment'' refers to the model trained without using any data augmentation.  
``Ours-less-datasets'' refers to the model trained without using latest datasets, namely FreiHAND and HO-3D. This model is trained with MTC, STB and RHD. These datasets are also used by previous methods.

%
%
Comparison between the variants of our model verifies that including various datasets and applying data augmentation are important in achieving better results. Inferring shape variations by estimating shape parameters is also helpful to improve the accuracy. 
%
%
Note that the results of ``Ours-less-datasets'' show that our model still achieves comparable performance on STB and RHD with the previous methods by using only limited datasets, and shows better 2D localization accuracy on in-the-wild MPII+NZSL dataset. 
These results demonstrate that our method takes advantage from both network design (including the training strategy) and larger training datasets.

%

We also qualitatively compare our method with previous work, as shown in Figure~\ref{fig:compare_sota_hand} and our supplementary video. The results indicate that our hand model can generate more precise 3D hand poses under challenging in-the-wild scenarios with occlusion, blur and low resolution.

\noindent \textbf{Ablation Study.}
We further examine two key designs used in training our hand module, the mixture of different datasets and data augmentation. 
The results for ablation study on the datasets are listed in Table~\ref{tab:ablation_dataset} and Figure~\ref{fig:ablation_dataset}. As expected, the results in Table~\ref{tab:ablation_dataset} shows that using more datasets will lead to better performance. 
We also show the examples of qualitative comparison in Figure~\ref{fig:ablation_dataset}. 
Similar to the conclusion from the quantitative study, the qualitative results show that incorporating more datasets can increase the models' generalization ability and generate more precise results for in-the-wild images.
In the figure, ``Subset-01'' means using the combination of datasets FreiHADN~\cite{Zimmermann:2019:FreiHAND} and HO-3D~\cite{Hampali:2019:HO3D}.
``Subset-02'' means using the combination of datasets: STB~\cite{Zhang:2017:Stereo}, RHD~\cite{Zimmermann:2017:Learning} and MTC~\cite{Xiang:2019:Monocular}. 
``Full set'' means using all the datasets.

%

The results for the ablation study on data augmentation are listed in Table~\ref{tab:ablation_data_augmentation} and Figure~\ref{fig:ablation_augment}.
The results in Table~\ref{tab:ablation_data_augmentation} demonstrate that applying data augmentation leads to better results. 
We also show qualitative results in Figure~\ref{fig:ablation_augment}, where, by adopting data augmentation, our models can generalize better to challenging scenarios including blur, challenging poses and occlusion.
In Figure~\ref{fig:ablation_augment}, ``No Augment'' refers to model trained without any data augmentation, and ``No Blur'' refers to model trained with all data augmentation strategies except motion blur augmentation.
 ``Full Augment'' refers to model trained with all data augmentation strategies.

\subsection{Integration Module Evaluation}
We qualitatively compare our method with previous whole body motion capture approaches, MTC~\cite{Xiang:2019:Monocular} and SMPLif-X~\cite{Pavlakos:2019:SMPLX}.
We also compare between two version of our model. 
We refer to the copy-and-paste integration and the optimization-based integration as ``Ours-CP'' and ``Ours-OP'', respectively
The results are shown in Figure~\ref{fig:compare_sota_body} and our supplementary videos, indicating that our method outperforms previous approaches in terms of both speed and accuracy. Notably, as shown in Table~\ref{tab:runtime}, our copy-and-paste method runs in two orders of magnitude faster speed than the alternative approaches, yet showing better 3D pose estimation quality. 





\section{Discussion}
\label{sec:Discussion}
We present FrankMacop, a fast motion capture system to estimate both 3D hand and 3D body motion from monocular inputs in the wild. 
We design the body and hand expert modules to produce compatible outputs for whole body motion capture. 
We present two integration strategies, copy-and-paste for faster speed and an optimization framework for better quality. 
The performance of our method has been demonstrated in-the-wild monocular videos. In particular, we also demonstrate our whole body motion capture system in a live demo at near real-time speed (9.5 fps), which is orders of magnitude faster than alternative methods. 
Our 3D hand pose estimation module outperforms previous state-of-the art on hand only methods in public benchmarks, and ours also can be used as a stand-alone monocular 3D hand pose estimator. 

Our method still suffered from a few limitations: 1. Hand pose estimation become erroneous if two hands are too close each other. 2. Bounding boxes are required to infer 3D body and hands. It would be an interesting future direction to mitigate these problems and extend the method to handle cases of multiple people interacting with each other, such as two people greeting with hand shaking.

\vspace {0.25cm}
\noindent \textbf{Acknowledgements.} We thank Yuting Ye for her helpful discussions and feedbacks. We also want to thank Xintao Wang for helping us in implementing motion blur augmentation.



{\small
\bibliographystyle{ieee_fullname}
\bibliography{hand_paper}
}

\end{document}